\documentclass[10pt,twocolumn,letterpaper]{article}

\usepackage[pagenumbers]{cvpr} %

\usepackage{algorithm}
\usepackage{algorithmic}
\usepackage{url}            %
\usepackage{booktabs}       %
\usepackage{amsfonts}       %
\usepackage{amssymb}     
\usepackage{amsmath}     
\usepackage{nicefrac}       %
\usepackage{microtype}      %
\usepackage{xcolor}         %
\usepackage{bm}     
\usepackage{xfrac} 
\usepackage{array}
\usepackage{makecell}
\usepackage{multirow}
\usepackage{enumitem}
\usepackage{color}
\usepackage{colortbl}

\definecolor{lightblue}{RGB}{122,46,246}
\definecolor{lightgreen}{RGB}{0,128,0}
\definecolor{lightred}{RGB}{226,3,9}
\definecolor{lightorange}{RGB}{225,165,0}
\definecolor{lightours}{RGB}{255,246,247}
\definecolor{FLOPs}{RGB}{248,203,173}

\newcommand{\PreserveBackslash}[1]{\let\temp=\\#1\let\\=\temp}
\newcolumntype{C}[1]{>{\PreserveBackslash\centering}p{#1}}
\newcolumntype{R}[1]{>{\PreserveBackslash\raggedleft}p{#1}}
\newcolumntype{L}[1]{>{\PreserveBackslash\raggedright}p{#1}}

\definecolor{cvprblue}{rgb}{0.21,0.49,0.74}
\usepackage[pagebackref,breaklinks,colorlinks,allcolors=cvprblue]{hyperref}

\def\paperID{*****} %
\def\confName{CVPR}
\def\confYear{2025}

\title{Stabilizing, Scaling \& Enhancing MeanFlow for Large-scale Diffusion Distillation}

\author{
    Xiao He\textsuperscript{1,2}, Yang Li\textsuperscript{2}, Peizhen Zhang\textsuperscript{2}, Songtao Liu \textsuperscript{2}, Zhao Zhong\textsuperscript{2} Nannan Wang\textsuperscript{1}\footnotemark[2], \\
    \textsuperscript{1}State Key Laboratory of Integrated Services Networks, Xidian University \\
    \textsuperscript{2}Tencent Hunyuan \\
    \tt\small nnwang@xidian.edu.cn
}

\begin{document}
\maketitle

\renewcommand{\thefootnote}{\fnsymbol{footnote}}
\footnotetext[2]{Corresponding author.}
\renewcommand{\thefootnote}{\arabic{footnote}}

\begin{abstract}
   Diffusion models exhibit remarkable generative capability, but their high latency limits practical deployment. Many studies have attempted to reduce sampling steps to accelerate inference. Among them, MeanFlow has attracted considerable attention due to its concise formulation and remarkable performance. Nevertheless, the instability of its optimization objective and the ``mean-seeking bias'' have limited its applicability to distill large-scale industrial models. To stabilize MeanFlow for distilling large-scale models, we first introduce a warm-up technique, in which the original differential solution of MeanFlow is replaced by a discrete solution. This design avoids training collapse caused by the MeanFlow target containing a stop-gradient term from an undertrained model. Once the model acquires a preliminary ability to fit the average velocity field, we switch the optimization objective back to the differential solution, enabling further refinement. Meanwhile, to alleviate the ``mean-seeking bias'' of MeanFlow under extremely few-step inference with complex target distributions, we incorporate trajectory distribution alignment as an auxiliary objective, encouraging the student model’s trajectory distribution to align more closely with that of the teacher model. Our proposed distillation framework achieves superior performance compared to existing distillation approaches when applied to the text-to-image (T2I) model FLUX.1-dev (up to 12B parameters). Furthermore, when extended to the 80B-parameter state-of-the-art (SOTA) T2I model HunyuanImage 3.0, our method continues to demonstrate robust generalization and strong performance.
\end{abstract}

\section{Introduction}
\label{sec:intro}
In recent years, diffusion and flow matching models~\cite{ho2020denoising,song2020score,liu2022flow,lipman2022flow} have emerged as the dominant frameworks in visual generation, replacing Generative Adversarial Networks (GANs)~\cite{goodfellow2020generative}. Their superior performance in terms of image quality, diversity, and scalability has driven the commercial adoption of generative technologies in areas such as image and video synthesis. However, diffusion models require dozens of iterative sampling steps, each involving a full neural network evaluation, leading to high computational costs and significant inference latency.

\begin{figure}
    \centering
    \includegraphics[width=1.0\linewidth]{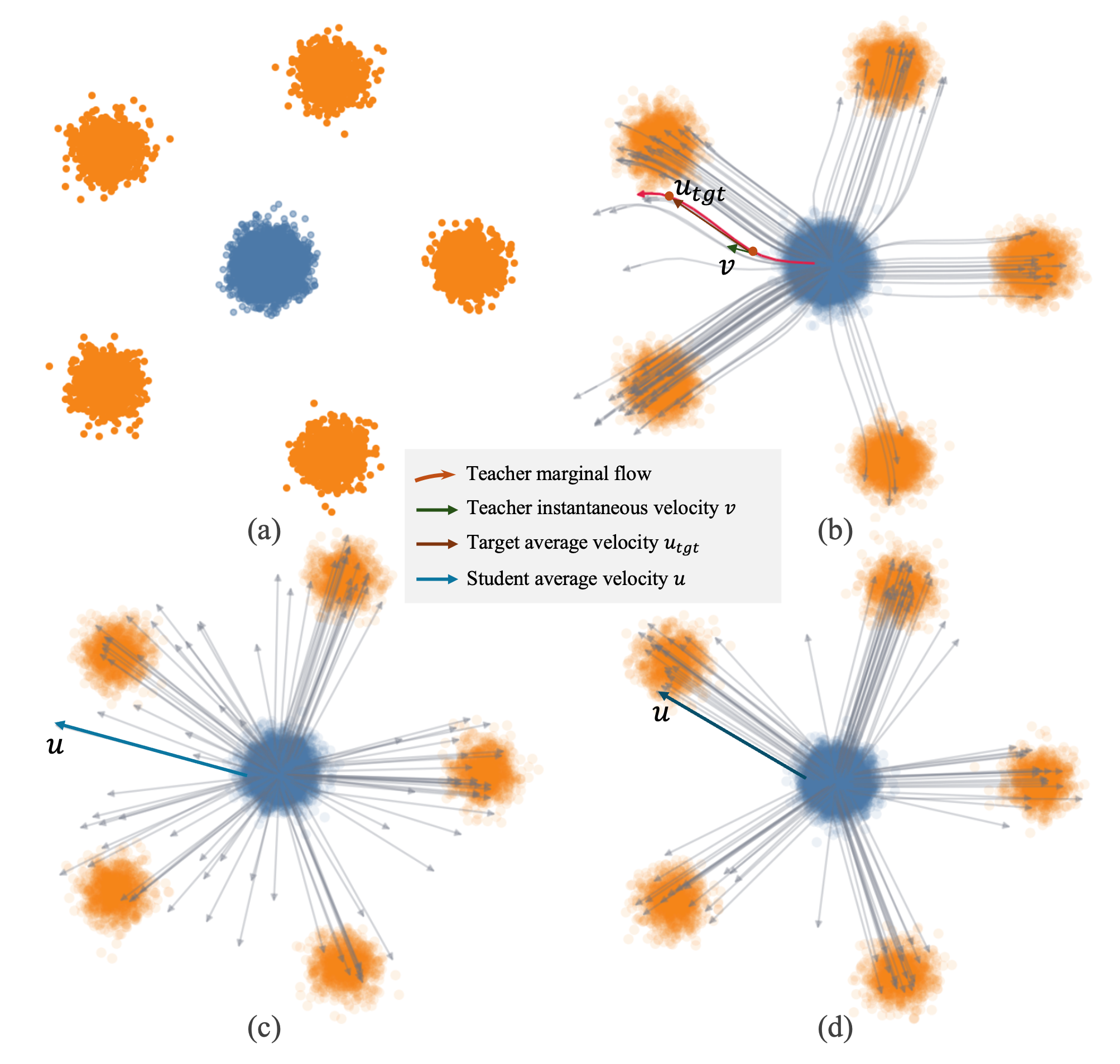}
    \caption{Illustration of the proposed method on toy Gaussian distributions. We adopt a tiny 2-layer MLP to learn the different velocity fields between source and target distributions to briefly introduce our method. (a) Source Gaussian noise in blue and target Gaussian mixture distribution in orange. (b) The field of instantaneous velocity, which is learned by a flow-based model viewed as the teacher in our framework. (c) One-step trajectory from the average velocity field, which is learned by the student by fitting the $u_{tgt}$ using the MeanFlow target. (d) Incorporating the trajectory distribution alignment objective effectively prevent the flow from the tendency to the mean of the target multimodal distribution.}
    \label{fig:toy_example}
\end{figure}

To reduce the inference overhead, existing studies employ distillation techniques to derive few-step diffusion models. These diffusion distillation methods can be broadly categorized into distribution-matching distillation~\cite{sauer2024fast,yin2024one,sauer2024adversarial,yin2024improved} and trajectory distillation~\cite{salimans2022progressive,song2023consistency,frans2024one}. The former typically employs score distillation~\cite{poole2022dreamfusion} or adversarial training to align the student’s output distribution with the target data distribution. However, this often disrupts the original diffusion sampling trajectory and limits multi-step sampling flexibility. The latter can achieve flexible multi-step sampling, but it will result in suboptimal image quality. 

Recently, continuous-time consistency models (sCM)~\cite{lu2024simplifying} have gained increasing attention for their solid theoretical foundations and elegant extensions, effectively eliminating the discretization errors inherent in discrete consistency models. Building upon this framework, IMM~\cite{zhou2025inductive} and MeanFlow~\cite{geng2025mean} incorporate both the current and target timesteps as conditioning inputs, enabling accurate predictions across varying inference budgets. Between them, MeanFlow has attracted particular attention for its simplicity and remarkable performance.

However, we observe that directly applying MeanFlow to distill large diffusion models fails to achieve satisfactory performance. Through extensive empirical analysis, we identify two primary causes: 1) \textit{Training instability} — caused by the MeanFlow target containing a stop-gradient term from an undertrained model and the numerical instability of Jacobian–vector product (JVP) computations~\cite{lu2024simplifying}. 2) ``Mean-seeking bias'' — when the target distribution is complex and the number of inference steps is extremely limited, the MeanFlow constraint on the expected velocity direction tends to drive predictions toward the mean of the target distribution, leading to blurry and overly smooth outputs\cite{lin2024sdxl}. The toy example in Fig.~\ref{fig:toy_example} (c) demonstrates this issue, where the average velocity, distilled using the MeanFlow regression target, tends to seek the mean direction of adjacent Gaussian peaks.

In this work, we introduce a framework to stablize, scale up, and enhance MeanFlow for large-scale difffusion distillation. 
First, to mitigate training instability, we derive a discrete solution of the MeanFlow target, and introduce a warm-up technique to equip the student with an initial ability to fit the average velocity field. Then, we adopt the differential objective to provide more precise supervision of the average velocity, which further refines the student model's output. With the robust initialization of the warm-up phase, the training can achieve fast and stable convergence. Moreover, to alleviate undesired short-cut learning that MeanFlow is proned to predict the mean of the multimodal distributions under limited inference steps, we incorporate a trajectory distribution alignment objective. It manages to enforce consistency between the endpoint distributions of the student and teacher trajectories over randomly sampled time intervals.
As shown in Fig.~\ref{fig:toy_example} (d), incorporating the trajectory distribution alignment objective effectively rectifes the flow away from the mean directions of the multimodal distribution. To reduce memory overhead and improve training efficiency, we developed JVP operators adapted for Flash Attention\cite{dao2023flashattention2} and various model parallel algorithms. Finally, we successfully scaled the MeanFlow distillation on two large-scale Text-2-Image (T2I) models: 12B-parameter FLUX.1-dev~\cite{flux2024} and 80B-parameter HunyuanImage 3.0~\cite{cao2025hunyuanimage}.
Compared with existing distillation methods, our approach achieves state-of-the-art performance and significant efficiency improvements on FLUX.1-dev. Furthermore, experiments on the 80B-parameter HunyuanImage 3.0~\cite{cao2025hunyuanimage} validate the effectiveness and generalization of our proposed method.

We highlight the contributions of this work as follows:
\begin{itemize}[topsep=0pt,parsep=0pt,leftmargin=18pt]
    \item We propose a novel distillation framework that, for the first time, enables the industrial-scale application of MeanFlow, providing a powerful and scalable solution for the distillation of large-scale models.
    \item We address the instability of MeanFlow training by introducing a warm-up strategy based on discrete MeanFlow objectives and incorporate trajectory distribution alignment to enhance the performance of models in few-step generation.
    \item Experiments on the popular FLUX.1-dev model demonstrate that our method outperforms existing few-step distillation baselines. Furthermore, evaluations on the 80B-parameter SOTA T2I model Hunyuan Image 3.0 validate the generalization and scalability of our approach.
\end{itemize}

\section{Related Works}
\label{sec:Related Works}

\subsection{Trajectory Distillation}
Diffusion models~\cite{ho2020denoising,song2020score} have been widely adopted owing to their outstanding generation quality. However, their iterative inference process—often requiring dozens of steps—introduces substantial latency, which limits practical deployment. Existing studies~\cite{salimans2022progressive,luo2023latent,song2023consistency,frans2024one} attempted to alleviate this issue by compressing the sampling trajectory through model distillation. A representative approach is trajectory distillation. Progressive distillation~\cite{salimans2022progressive} was initially proposed to gradually compress diffusion models, but its multi-stage distillation process leads to error accumulation. Subsequently, the consistency model (CM)~\cite{song2023consistency} maps any trajectory point directly to the diffusion starting point, significantly accelerating the sampling process. Building upon this idea, numerous follow-up works~\cite{luo2023latent,frans2024one,kim2023consistency} have been developed based on the concept of consistency. Nevertheless, its multi-stage training and discretization errors result in complex optimization and limit its ultimate performance. Recently, continuous-time CM (sCM)~\cite{lu2024simplifying} has been proposed and demonstrated to be effective on academic image benchmarks. Building on this, MeanFlow~\cite{geng2025mean} introduced the concept of the average velocity field, achieving remarkable performance on academic benchmarks. Despite these advancements, instability during training and compatibility issues with JVP computations hinder the applicability of such methods to large-scale, industrial-level diffusion models. Although several previous works~\cite{lu2024simplifying,chen2025sana,zheng2025large} have explored stabilized training strategies for differential solution of sCM, these approaches often require modifications to time embeddings or introduce approximation errors, rendering them suboptimal for practical large-scale deployment. Concurrent with our work, CMT~\cite{hu2025cmt} explored training-stabilization techniques similar to ours, but its effectiveness was not validated on large-scale models. Moreover, our framework further incorporates trajectory distribution alignment, which enhances the model’s few-step generation capability.

\subsection{Distribution Matching Distillation}
Unlike trajectory distillation, which aims to learn the teacher model’s sampling trajectories, distribution-matching distillation methods~\cite{sauer2024adversarial,sauer2024fast,yin2024one,yin2024improved} focus on aligning the output at the distribution level. Specifically, it accelerates diffusion sampling by matching the student model’s output distribution to either the teacher’s prior or the target data distribution. Adversarial Diffusion Distilation (ADD)~\cite{sauer2024adversarial,sauer2024fast} achieves this through score distillation sampling (SDS)~\cite{poole2022dreamfusion} combined with adversarial training. However, due to the inherent limitations of SDS, the generated images often appear blurry or oversaturated. To address this issue, a series of subsequent works~\cite{yin2024one,yin2024improved,he2024one,zhou2024score} have improved upon the original score distillation framework and achieved promising results using few-step sampling in various visual generation tasks.Yet,these methods train at fixed time steps and align distributions in a single forward pass, which disrupts the model’s sampling trajectory and limits multi-step flexibility. Trajectory Distribution Matching (TDM)~\cite{luo2025learning} mitigates this issue by dividing the diffusion trajectory into segments and performing distribution matching within each segment, improving sampling flexibility.  In contrast, our distillation framework organically integrates the principles of trajectory distillation and distribution matching, avoiding the need for manual trajectory segmentation as required by TDM, and without relying on a specific form of distribution-matching objective.

\section{Method}
\label{sec:method}

\subsection{Preliminary}
In this section, we briefly review the most related works and summarize the notations used in this paper.

\textbf{Flow Matching.}
Diffusion models~\cite{ho2020denoising,song2020score} gradually add noise to clean data and train neural networks to reverse this process. This denoising process can be formulated as solving the stochastic differential equation (SDE) or their deterministic counterparts, probability flow ordinary differential equation (PF-ODE)~\cite{song2020score}. 
Flow Matching ~\cite{liu2022flow,lipman2022flow} further extends this formulation by modeling the velocity field along the flow paths between two probability distributions. 
Formally, given a data sample $x \sim p_{data}$ and a noise sample $\epsilon \sim p_{prior}$, a flow path is defined as $z_t = a_tx + b_t\epsilon$, $\forall t \in [0, 1]$. $a_t$, $b_t$ are the predefined scheduling functions, typically $a_t = 1-t$ and $b_t = t$. The time derivative of the flow path is defined as the conditional velocity $v(z_t | x)$.
The goal of Flow Matching is to fit the marginal velocity $v(z_t, t)$, which is the expectation of the conditional velocity over all $(x, \epsilon)$ pairs consistent with $z_t$. To circumvent the intractability of directly learning marginal velocities, Conditional Flow Matching (CFM)~\cite{tong2023conditional} trains a network $v_{\phi}$ to match the tractable conditional velocity along flow paths using the objective:
\begin{equation}
\mathcal{L}_{CFM}(\theta) = \mathbb{E}_{t,x,\epsilon}||v_\phi (z_t,t)-v(z_t | x)||^2.
\label{eq:ode}
\end{equation}
It can be shown that minimizing this loss is equivalent to fitting true marginal velocity field, and enables efficient simulation-free training.

With the parametric marginal velocity $v_{\phi} (z_t,t)$, the sampling process can be achieved by solving the following ODE using Euler sampling over small discrete time intervals,
\begin{equation}
\frac{d}{dt}z_t = v_\phi (z_t,t).
\label{eq:ode}
\end{equation}


\textbf{MeanFlow.}
The core idea of MeanFlow~\cite{geng2025mean} is to introduce a new field representing average velocity, whereas the velocity modeled in Flow Matching represents the instantaneous velocity.
For any two timesteps $t$ and $r$, the average velocity $u(z_t,r,t)$ is defined as the integral of the instantaneous velocity over a given time interval $[r, t]$,

\begin{equation}
u(z_t,r,t) = \frac{1}{t-r}\int_{r}^{t} v(z_\tau,\tau) d\tau.
\label{eq:avg_v}
\end{equation}
To derive a trainable formulation, MeanFlow transforms the integrated form into the following differential form: 
\begin{equation}
 u(z_t,r,t) = v(z_t, t)-(t-r)\frac{d}{dt}u(z_t,r,t).
\label{eq:u_tgt}
\end{equation}
This equation is termed the ``MeanFlow Identity'' and serves as the core supervision signal for efficiently training one/few-step generative model.


\begin{figure*}
    \centering
    \includegraphics[width=1.0\linewidth]{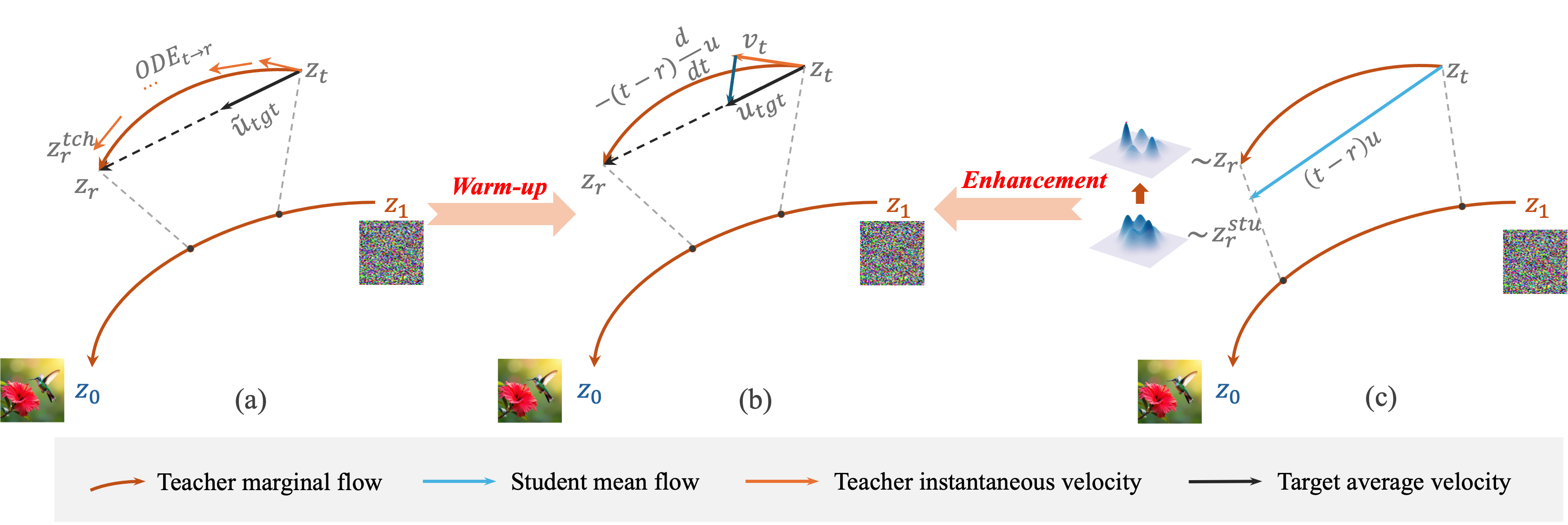}
    \caption{Illustration of the proposed distillation framework. (a) \textbf{Warm-up}: achieves a robust initialization of average velocity field by fitting $\tilde{u}_{tgt}$ derived from a discrete approximation. (b) \textbf{Fine-tuning}: refines the student capbility via the more accurate supervision signals $u_t$. (c) \textbf{Enhancement}: incorporates the trajectory distribution alignment objective to prevent \textit{mean-seeking bias} leading to superior few-step performance.}
    \label{fig:arch}
\end{figure*}

\textbf{Distribution Matching Distillation.}
Distribution-matching distillation primarily leverages score distillation~\cite{poole2022dreamfusion} or adversarial training~\cite{sauer2024adversarial} to optimize the student model. DMD~\cite{yin2024one} adopts a variational score distillation objective~\cite{wang2023prolificdreamer}, encouraging the student-generated images to follow the prior distribution of teacher. LADD~\cite{sauer2024fast} aligns the diffusion distribution between the student’s outputs and real data through adversarial learning. Although these methods achieve promising results, they often overlook the intermediate steps of the diffusion trajectory, thereby disrupting the original sampling dynamics of the diffusion model.

\subsection{Overview and Analysis}


This section presents the proposed distillation framework and analyzes the key challenges. 
Given a pretrained flow-based model $G_\phi$ with its marginal velocity field $v_\phi (z_t,t)$, we aim to distill a few-step student generator $G_\theta$. To achieve few-step generation, we equip the student with an additional timestep input $r$, and learn an average velocity field $u_\theta(z_t,r,t)$ by minimizing the following objective: 

\begin{equation}
\mathcal{L}_\theta = \mathbb{E}||u_\theta(z_t,r,t) - sg(u_{tgt})||_2^2,
\label{eq:meanflow_jvp}
\end{equation}
where $u_{tgt}$ serves as the regression target, and $sg$ represents the stop-gradient operator.
MeanFlow provides an effective differential solution to calculate the target. Differently,  we replace the $v(z_t,t)$ with the known marginal velocity $v_\phi (z_t,t)$ provided by the pretrained teacher, and formulate $u_{tgt}$ as:
\begin{equation}
u_{tgt} = v_{\phi}(z_t,t)-(t-r)\frac{d}{dt}u_{\theta}(z_t,r,t).
\label{eq:u_tgt}
\end{equation}

However, directly apply the above method to large-scale diffusion distillation has several challenges:

1. The $u_{tgt}$ relies on the stop‑gradient of $u_\theta$ produced by the undertrained student generator. This results in poor supervision. In addition, the time-derivative term $\frac{d}{dt}u_{\theta}$ suffers from severe numerical instability in optimization \cite{lu2024simplifying}. These two factors severely hinder convergence in the early stages of distillation training, particularly for large-scale diffusion models.

2. Although the Mean Squared Error (MSE) loss in Eq.~\ref{eq:meanflow_jvp} can effectively guide the student to learn a reasonable average velocity field, it may introduces a \textit{mean-seeking bias}\cite{lin2024sdxl}, which drives the estimated velocity direction toward the mean of the target distribution under few-step inference (as shown in Fig.~\ref{fig:toy_example}). As the target distribution scales in complexity (e.g., high-resolution image generation), the issue becomes increasingly severe.

3. For implementation, the time-derivative term $\frac{d}{dt}u_{\theta}$ is expanded in terms of partial derivatives $\frac{d z_{t}}{d t} \partial_{z} {u_\theta}+\frac{d r}{d t} \partial_{r} {u_\theta}+\frac{d t}{d t} \partial_{t} {u_\theta}$,  and computed by the JVP. However, The JVP operator is not well-suited for large-scale model training infrastructure.

To alleviate these challenges, we propose a systematic distillation framework that effectively scales MeanFlow for large-scale diffusion distillation. The details are presented in the following sections.

\begin{figure}
    \centering
    \includegraphics[width=1.0\linewidth]{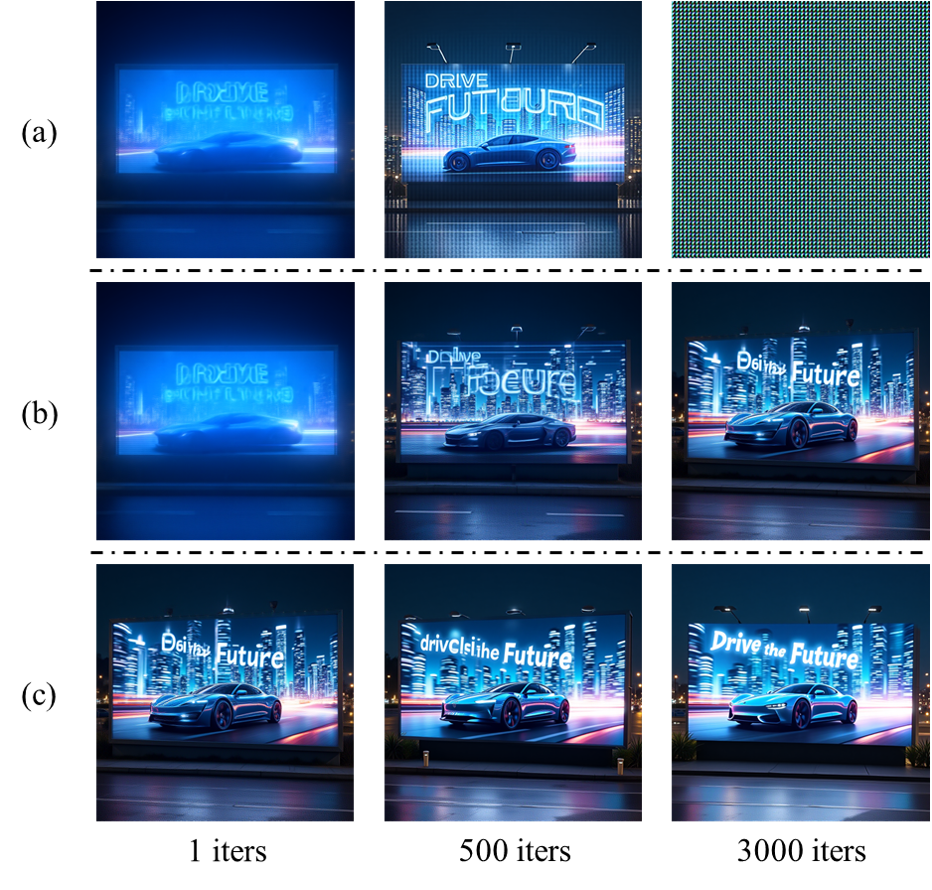}
    \caption{Visualization of the student outputs under different training strategies and iterations.(a) Directly training the student with the differential objective. (b) Training the student model with the discrete objective. (c) Fine-tuning the student with the differential objective after warm-up training. }
    \label{fig:train_cmp}
\end{figure}

\subsection{Warm-up with Discrete Objective}

In order to mitigate the convergence and training instability issues, 
we initially train the student with a target that neither depends on the student’s predictions nor involves unstable time-derivative computations.
Specifically, we return to the fundamental definition of the average velocity (Eq.~\ref{eq:avg_v}) and compute the target average velocity through discrete approximation. 
As shown in Fig.~\ref{fig:arch}(a), given a sampled starting point $z_t$ and a timestep interval $[r, t]$, we can obtain the instantaneous velocity from the teacher model and get the point along the trajectory via the ODE solver~\cite{song2020score}. Once the trajectory endpoint $z_r^{tch}$ is obtained, the target average velocity can be calculated as follows:


\begin{equation}
\tilde{u}_{tgt} = \frac{z_t - z_r^{tch}}{t-r},
\label{eq:u_tgt_dis}
\end{equation}

\begin{equation}
z_r^{tch} = \text{ODE}(v_\phi(z_\tau,\tau))|_{\tau=t\rightarrow r}.
\label{eq:z_r_tch}
\end{equation}

As shown in the Fig.~\ref{fig:train_cmp} (a), when directly training the student with the differential objective, the quality of the generated image fails to improve and eventually collapses at 3,000 iterations. In contrast, the discrete objective leads to steady improvement and stable convergence, as seen in Fig.~\ref{fig:train_cmp} (b). The loss curves shown in Fig.~\ref{fig:loss_ablation} also explain the significant difference in convergence between the two configurations. However, the discrete objective causes imprecise supervision during training, which will limit the upper bond of the student performance. Thus, we just employ this as a warm-up phase to establish a robust initialization for the subsequent optimization of the differential objective.


\subsection{Fine-tuning with Differential Objective}
After the warm-up, the student model has gained a preliminary ability to estimate the average velocity and can relatively stably compute the differential solution. Thus, we transition to using Eq.~\ref{eq:u_tgt} as the target average velocity to continuously train the model (as shown in Fig.~\ref{fig:arch} (b)). The differential objective provides more precise supervision signals, and further improves the capacity to predict the average velocity. As shown in Fig.~\ref{fig:train_cmp} (c), the fine-tuning effectively improves the sharpness and richness of details in the generated images.


\subsection{Trajectory Distribution Alignment}

In this section, based on the MeanFlow optimization objective, we introduce a trajectory distribution alignment objective as a regularization term to mitigate the ``mean-seeking bias'' of MeanFlow under few-step inference. As illustrated in Fig.~\ref{fig:arch} (c), this objective imposes additional constraints on the endpoints of trajectories sampled by the student model, encouraging close alignment between the trajectory endpoint distributions of the student and teacher models. The overall optimization objective for this stage can therefore be formulated as:

\begin{equation}
\mathcal{L}_\theta = \mathbb{E}||u_\theta(z_t,r,t) - sg(u_{tgt})||_2^2 + \lambda \mathcal{L}_{TDA},
\label{eq:loss_second}
\end{equation}
where $u_{tgt}$ is computed according to Eq.~\ref{eq:meanflow_jvp}, 

Common techniques for distribution alignment include score distillation and adversarial training. In our implementation, we employ a GAN-based approach to achieve distribution alignment. The distribution of trajectory endpoints sampled by the teacher model over multiple steps is treated as the true distribution (here, $z_r \approx z_r^{tch}$), while the distribution of trajectory endpoints sampled by the student model serves as the fake distribution for the discriminator (i.e., $\mathcal{L}_{TDA} = \mathcal{L}_{adv}(z_r^{stu},z_r^{tch})$). Specifically, after sampling noise $z_1$, data $z_0$, and timesteps $t$ and $r$, noisy latent $z_t$ is obtained according to the noise schedule. This latent is then fed into both the teacher and student models. The teacher model obtains the target latent $z_r^{tch}$ through ODE-based sampling. In contrast, The student model predicts the mean velocity $u_{\theta}$ within this time interval and samples the corresponding latent $z_r^{stu}$, which can be formulated as follows:

\begin{equation}
z_r^{stu} = z_t - (t-r)u_{\theta}(z_t,r,t).
\label{eq:z_r}
\end{equation}

Finally, we perform adversarial training on these two endpoints to encourage the student model’s distribution to align closely with that of the teacher model. As shown in the toy experiment in Fig.~\ref{fig:toy_example}(d), incorporating trajectory distribution alignment leads to a more accurate predicted average velocity direction, effectively guiding it toward the real target distribution.

\begin{table*}[t]
    \centering
    \caption{Quantitative comparisons of our method and competitors on GenEval benchmark.}
    \label{tab:geneval}
    \small
    \begin{tabular}{cc|ccccccc}
       \toprule
       \multirow{2}*{Methods} & \multirow{2}*{NFE} &Single  &Two &\multirow{2}*{Counting} & \multirow{2}*{Colors} & \multirow{2}*{Position} &Color  &\multirow{2}*{Overall} \\
       & &Object &Object &  & & &Attribution & \\
        \midrule
          FLUX.1-dev\,\cite{flux2024}  & 50  & 0.98 & 0.81 & 0.74 & 0.79 &0.22 & 0.45 &0.66  \\
          FLUX Turbo\,\cite{FLUX-turbo}  &8 & 0.97 & 0.77 & 0.71  & 0.78 & 0.24 & 0.40  & 0.64 \\
          Hyper-Flux\,\cite{ren2024hyper} & 8 & 0.98 & 0.81 & 0.69 & \textbf{0.81} & 0.20 & 0.52 & 0.67 \\
          FLUX.1-schnell\,\cite{flux2024}  & 4  & \textbf{1.00 }& \textbf{0.87} & 0.57 & 0.75 & \textbf{0.28} & 0.51 & 0.66 \\
          $\pi$-Flow\,\cite{chen2025pi} & 4 & 0.98 &0.82 & 0.71 & 0.80 & 0.23 & 0.52 &0.68 \\
          Ours& 4 & 0.99 & 0.81  &  \textbf{0.79} & 0.79 &  0.21& \textbf{0.52} &\textbf{0.69}  \\
       \bottomrule
    \end{tabular} 
 \end{table*} 
\begin{figure*}
    \centering
    \includegraphics[width=1.0\linewidth]{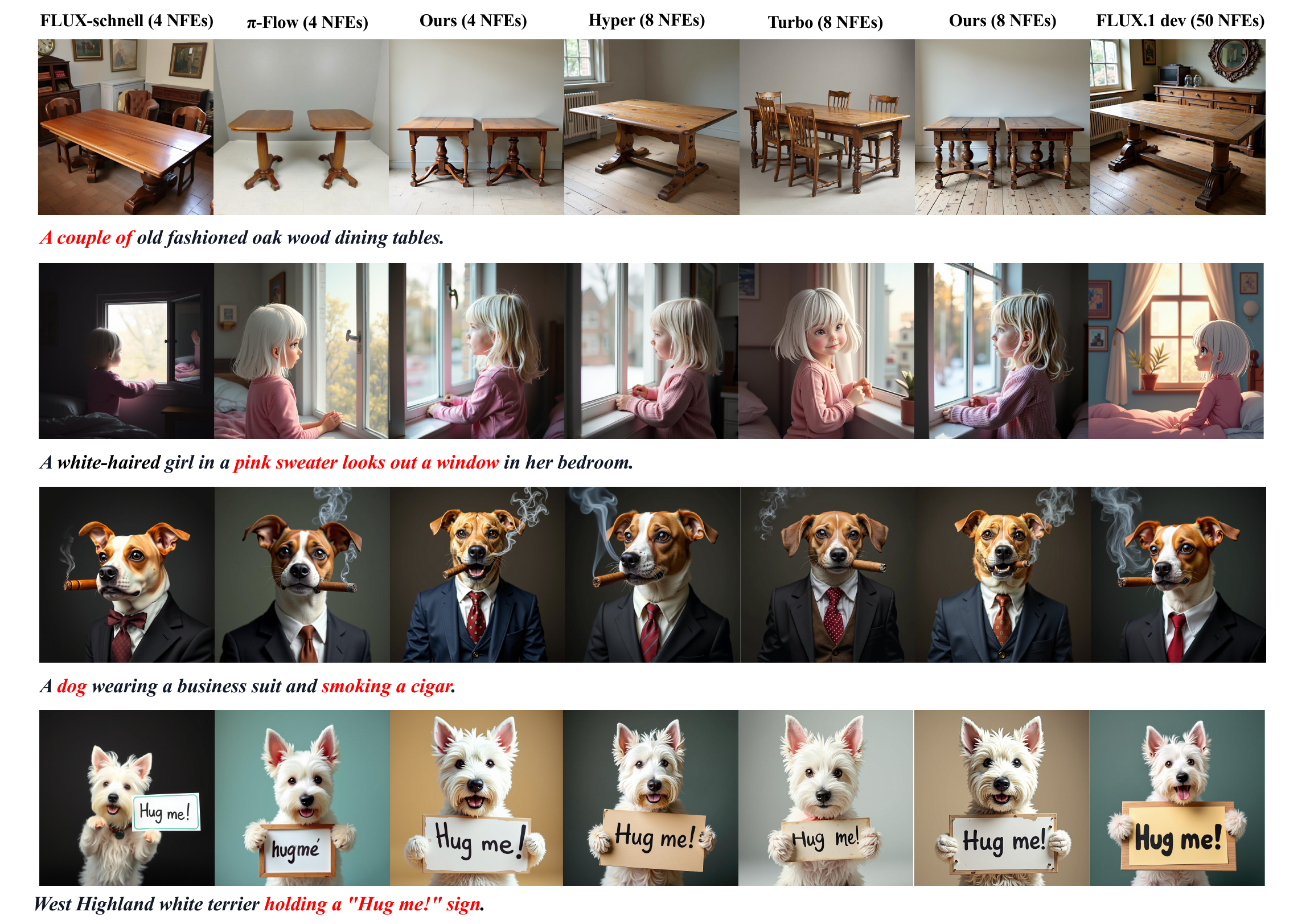}
    \caption{Qualitative comparisons of our method with competitors. NFE denotes the number of function (network) evaluations.}
    \label{fig:flux_cmp}
\end{figure*}
\section{Experiments}
\label{sec:Experiments}
\subsection{Experiment Setting}

\textbf{Baselines.}
We validate the effectiveness of our proposed methods on the challenging text-to-image generation task. In terms of generalization and scaling-up ability, we perform the studies on top of a publicly yet strong enough baseline (\textit{i.e.}, FLUX\,\cite{flux2024}) and then examine our efficacy upon the SOTA model (\textit{i.e.}, HunyuanImage3.0\,\cite{cao2025hunyuanimage}, 1st place in LMArena\,\cite{lmarena-t2i}). 

\noindent\textbf{Benchmarks.}
We choose the prevalent GenEval\,\cite{ghosh2023geneval} and HPSv2.1\,\cite{wu2023human} Benchmarks for assessing model capability. The former one reveals the ability in prompt adherence about general attributes, quantities and colors, etc. The latter one is more human preference-proned. Specifically, text-image alignment (\textit{CLIP Score\,\cite{hessel2021clipscore}}) is also given \textit{w.r.t.} on four visual categories.

\begin{table*}[t]
    \centering
    \caption{Quantitative comparisons of our method and competitors on HPSv2.1 benchmark.}
    \label{tab:hpsv2}
    \small
    \begin{tabular}{cc|cccccc}
       \toprule
        \multirow{2}*{Methods} &  \multirow{2}*{Steps}  & \multicolumn{5}{|c|}{HPS$\uparrow$}  & \multirow{2}*{CLIP$\uparrow$} \\
        &    & \multicolumn{1}{|c}{Animation}  & Concept-Art & Painting & Photo & \multicolumn{1}{c|}{Average}   & \\
          \midrule
          FLUX.1-dev\,\cite{flux2024}  & 50   & 32.50 & 31.06 & 31.51 & 29.96 & 31.26 & 0.327  \\
          FLUX Turbo\,\cite{FLUX-turbo}  &8 & 32.94 & 31.23& 31.78 & 30.01 & 31.49 & 0.328  \\
          Hyper-Flux\,\cite{ren2024hyper} & 8 & 33.41 & 31.56 & 31.83 & 30.31 & 31.78 & 0.327   \\
          FLUX.1-schnell\,\cite{flux2024}  & 4  & 30.49 & 29.47 & 30.18 & 28.86 & 29.75 & \textbf{0.333}  \\
          $\pi$-Flow\,\cite{chen2025pi} & 4 & 32.59 & 31.10 & 31.49 & 30.11 & 31.32 & 0.328  \\
          Ours& 4 & \textbf{33.43} & \textbf{31.72}  &\textbf{31.96}  &\textbf{30.65}  & \textbf{31.94} & 0.328  \\
       \bottomrule
    \end{tabular} 
 \end{table*}

\begin{figure*}
    \centering
    \includegraphics[width=1.0\linewidth]{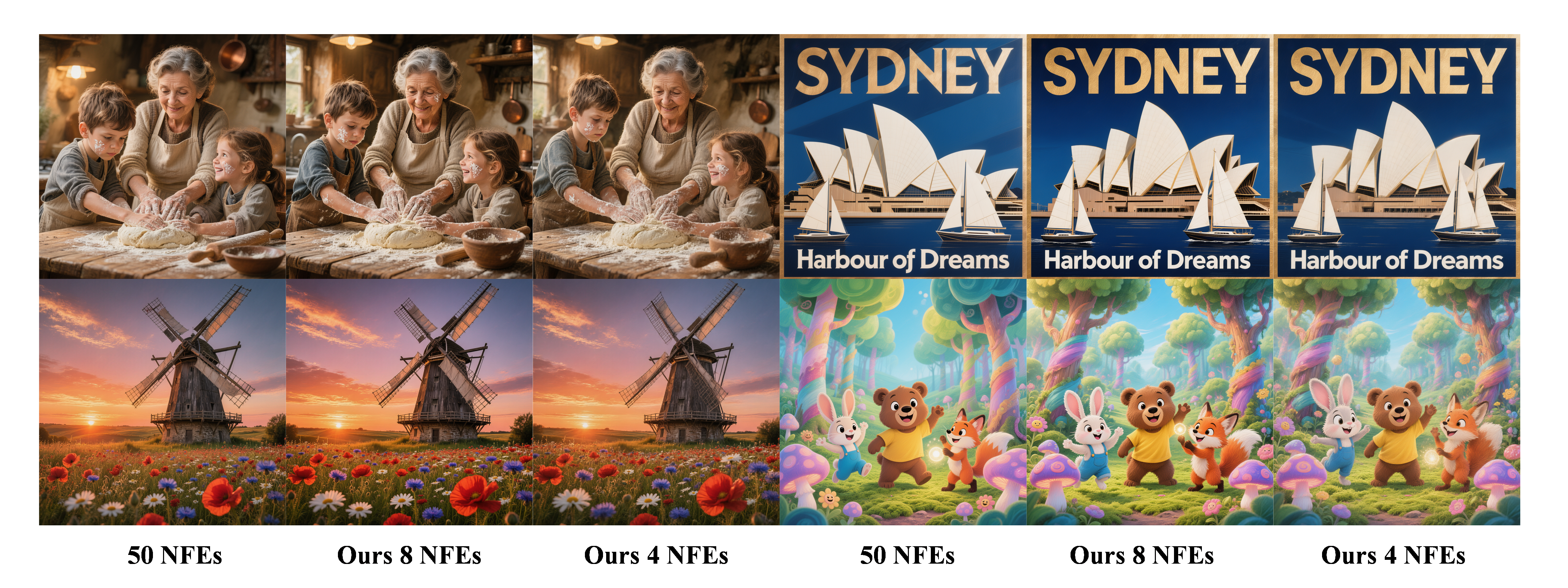}
    \caption{Visual comparison between the few-step generation results of our distilled model and the 50-NFE results of the original HunyuanImage 3.0 model.}
    \label{fig:HY3.0_cmp}
    \vspace{-2mm}
\end{figure*}

\noindent\textbf{Instantiation.}
Our experiments adopt a two-stage training strategy. In the warm-up stage, the student model is trained for approximately 3K iterations using the average velocity computed by Eq.~\ref{eq:u_tgt_dis} as the target. In the second stage, we continue optimizing the student model based on this objective. Furthermore, during the this training phase, we activate the GAN loss and set the balancing coefficient $\lambda$ in Eq.~\ref{eq:loss_second} to $10$ to balance the two loss terms. The entire training process spans approximately 6K iterations in total. To enable large-scale model training, we develop an efficient JVP operator of Flash Attention, and suite JVP to a series of model‑parallel regimes.

\subsection{Comparisons}
 
\textbf{Comparison upon strong baseline.}
We select the most competitive FLUX-based few-step distilled models for comparison, including: (1) Hyper-Flux (an 8-step lora upon FLUX.1-dev mentioned in Hyper-SD\,\cite{ren2024hyper})\, trained via consistency distillation and reinforcement learning, (2) Flux Turbo\,\cite{FLUX-turbo} trained with adversarial learning, (3) Official Flux.1-Schnell\,\cite{flux2024} trained through latent diffusion distillation and (4) $\pi$-Flow\,\cite{chen2025pi}, a policy-based flow model trained via imitation distillation. For fair study, we apply our proposed method on the publicly available FLUX.1-dev model to achieve a few-step distilled version and compare to the aforementioned ones. Table\,\ref{tab:geneval} exhibit the results on \textit{GenEval}. The few-step-distilled model by our method not only surpasses all previous 4-step FLUX-based counterparts, but also performs better than the 8-step variants. What's more, it also outperforms the 50-step teacher while achieving 12.5$\times$ steps reduction. Notably, the counting ability of our model is apparently better than that of the others'. Table\,\ref{tab:hpsv2} shows the results on \textit{HPS} benchmark. We adopt the newest \textit{HPSv2.1} scoring manner for comparison. We could observe that our method is superior to the previous ones, upon all four categories. This demonstrates the generalization ability towards diverse prompting scenarios. In terms of CLIP scores, all methods achieve relatively similar performance, indicating that the generated images maintain strong semantic correspondence with the prompt.

In addition, Fig.~\ref{fig:flux_cmp} presents a visual comparison among different methods. As shown in the figure, our approach demonstrates several clear advantages: 1. Stronger prompt adherence — as illustrated in the first row of Fig.~\ref{fig:flux_cmp}, only our method successfully generates a couple of tables as specified in the prompt, whereas other methods produce only one. This superior attribute alignment is further validated by the quantitative results in Table~\ref{tab:geneval}. 2. Better structural consistency — when generating creative or uncommon scenes (e.g.,``a dog smokin''), our method produces plausible and coherent structures without introducing distortions or noticeable artifacts. 3.Finer detail preservation — our model effectively reproduces high-frequency textures such as the sweater fabric and dog fur in Fig.~\ref{fig:flux_cmp}. 4. Flexible sampling capability — our method supports adaptive adjustment of sampling steps within a certain range after training, enabling efficient trade-offs between inference cost and visual quality under varying computational budgets.

\noindent\textbf{Scaling-up ability towards SOTA.}
We further apply our distillation framework to the SOTA T2I model HunyuanImage 3.0 to assess its generalizability. Fig.~\ref{fig:HY3.0_cmp} presents a visual comparison between the distilled model’s 4-NFE and 8-NFE results and the original model’s 50-NFE results. As shown in the figure, the 8-NFE results of the student model are nearly indistinguishable from those of the original model, while the 4-NFE results exhibit only minor deviations in fine details. These observations strongly demonstrate the effectiveness and generalizability of our method.

\begin{figure}
    \centering
    \includegraphics[width=1.0\linewidth]{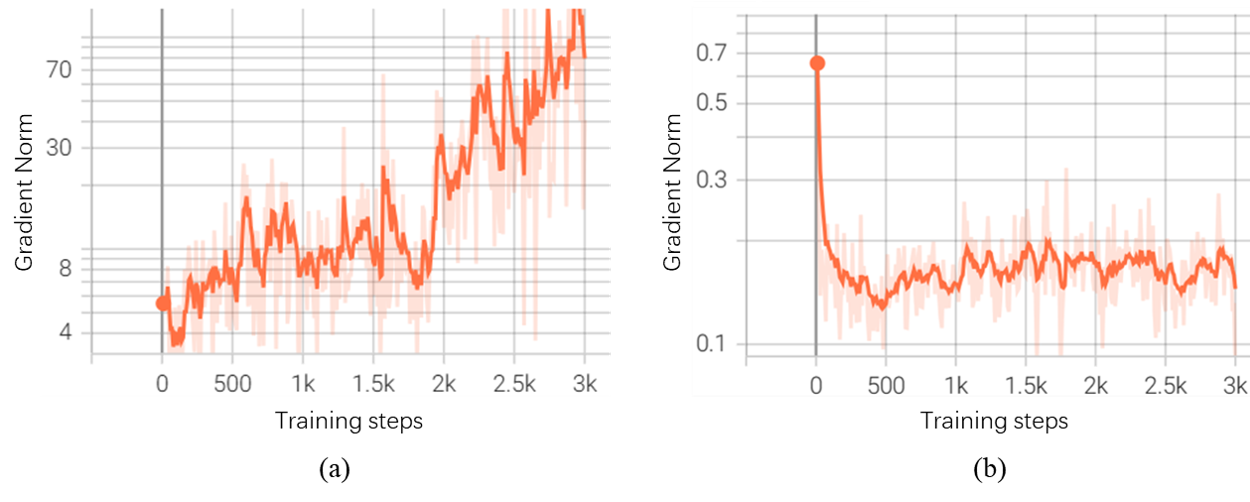}
    \caption{Loss curves trained under different optimization objectives. (a) Training with the differential objective. (b) Training with the discrete objective.}
    \label{fig:loss_ablation}
\end{figure}

\begin{table}[t]
    \centering
    \caption{Comparisons of variants on HPSv2.1 and GenEval benchmarks under 4-NFE configuration.}
    \label{tab:ablation}
    \small
    \begin{tabular}{c|cc|c}
       \toprule
       Setting & HPS $\uparrow$ &CLIP $\uparrow$ &GenEval $\uparrow$ \\
        \midrule
          w/o Warm up    & N/A & N/A   & N/A \\
          w/o  TDA   & 30.25 &  0.326 & 0.62 \\
          Ours (VSD-based TDA) & 30.66 & \textbf{0.33} &  0.64   \\
          Ours & \textbf{31.94} & 0.328   &\textbf{0.69}  \\
       \bottomrule
    \end{tabular} 
 \end{table}

\subsection{Ablation study}
In this section, we investigate the impact of the key components in our proposed distillation framework, including the warming-up training with the discrete meanflow objective and the trajectory distribution alignment.

\noindent\textbf{Warm-up training.} We have analyzed the instability of MeanFlow’s differential solution and introduced a warm-up strategy to stabilizes student model training. To validate its effectiveness, we compare the loss convergence behaviors of models trained with the differential and discrete MeanFlow objectives, respectively. As shown in Fig.~\ref{fig:loss_ablation}, directly optimizing the model using the differential solution of MeanFlow leads to training collapse, whereas replacing the target with the discrete integral formulation results in stable and convergent optimization. Moreover, the visualization results in Fig.~\ref{fig:train_cmp} demonstrate that, after the warm-up training phase, the student model obtains a robust initialization that facilitates subsequent differential MeanFlow optimization and enhances its ability to fit the average velocity field. The results demonstrate the importance of the proposed warm-up strategy for achieving stable training.

\begin{figure}
    \centering
    \includegraphics[width=1.0\linewidth]{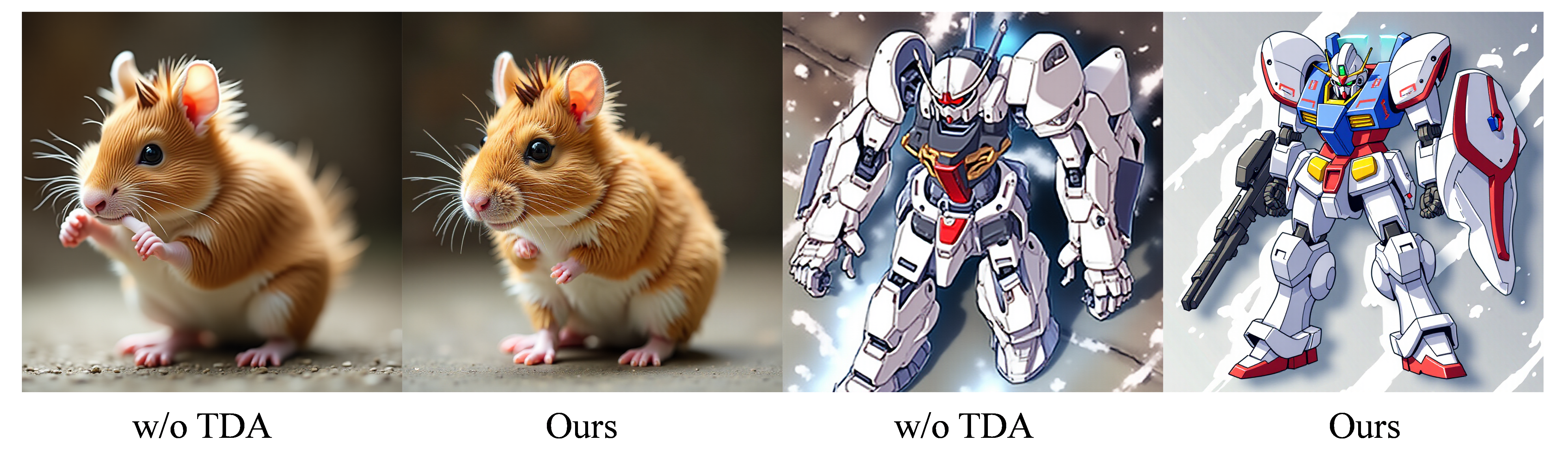}
    \caption{Impact of trajectory distribution alignment on model.}
    \label{fig:TDA_cmp}
    \vspace{-2mm}
\end{figure}

\noindent\textbf{Trajectory Distribution Alignment.}
To evaluate the impact of the proposed trajectory distribution alignment on model performance, we designed two variants: one without trajectory distribution alignment (denoted as w/o TDA), and another that adopts variational score distillation\,\cite{wang2023prolificdreamer} (VSD) for trajectory alignment (denoted as VSD-based TDA). In VSD-based TDA, the endpoint samples $z_r^{stu}$ generated by the student model are fed into the teacher model for prediction, while another set is predicted by a fake score network, and their outputs are subsequently aligned~\cite{yin2024one}. As shown in Table~\ref{tab:ablation}, removing trajectory distribution alignment leads to a substantial drop in human preference scores. Moreover, Fig.~\ref{fig:TDA_cmp} illustrates that omitting trajectory distribution alignment results in noticeably degraded image sharpness and detail. Incorporating trajectory distribution alignment—regardless of the specific formulation—significantly enhances model performance, underscoring its critical role within our distillation framework.

\section{Conclusion}
\label{sec:conclusion}

In this work, we propose a novel MeanFlow-based distillation framework that successfully scales MeanFlow to distill large-scale diffusion models. Specifically, by introducing a discrete MeanFlow objective for warm-up training, we successfully mitigate the instability issues encountered when applying it to distill large-scale models. Furthermore, we integrate trajectory distribution alignment into the MeanFlow optimization objective, alleviating the ``mean-seeking bias'' of MeanFlow under limited inference steps with complex target distributions. Extensive experiments demonstrate that our distillation framework achieves competitive text-to-image generation quality while reducing inference steps by up to $12\times$ compared to the teacher models.

{
    \small
    \bibliographystyle{ieeenat_fullname}
    \bibliography{main}
}

\end{document}